\documentclass[letterpaper, 10 pt, conference]{ieeeconf}  

\IEEEoverridecommandlockouts                              

\usepackage{graphicx}
\usepackage{amsmath} 
\usepackage{authblk}
\usepackage{pifont}
\newcommand{\cmark}{\ding{51}}%
\newcommand{\xmark}{\ding{55}}%
\usepackage[colorlinks = true,
            linkcolor = blue,
            urlcolor  = blue,
            citecolor = blue,
            anchorcolor = blue]{hyperref}
            
\usepackage[normalem]{ulem}
\useunder{\uline}{\ul}{}

\title{\LARGE \bf
Robust Human Motion Forecasting using Transformer-based Model
}

\author[1]{Esteve Valls Mascaro$^{*}$ \thanks{$^{*}$These authors equally contributed to this work.}

\thanks{This work is funded by Marie Sklodowska-Curie Action Horizon 2020 (Grant agreement No. 955778) for project 'Personalized Robotics as Service Oriented Applications' (PERSEO).}
        }

\author[2]{Shuo Ma$^{*}$
}
\author[3]{Hyemin Ahn }
\author[1,4]{Dongheui Lee }

\affil[1]{\small{Autonomous Systems, Technische Universität Wien (TU Wien), Austria}}
\affil[2]{Human-centered Assisitve Robotics, Technische Universität München (TUM), Germany}
\affil[3]{Artificial Intelligence Graduate School, Ulsan National Institute of Science and Technology (UNIST), Korea}
\affil[4]{Institute of Robotics and Mechatronics, German Aerospace Center (DLR), Germany}

\begin{document}

\maketitle
\thispagestyle{empty}
\pagestyle{empty}

\begin{abstract}

Comprehending human motion is a fundamental challenge for developing Human-Robot Collaborative applications. Computer vision researchers have addressed this field by only focusing on reducing error in predictions, but not taking into account the requirements to facilitate its implementation in robots. In this paper, we propose a new model based on Transformer that simultaneously deals with the real time 3D human motion forecasting in the short and long term. Our 2-Channel Transformer (2CH-TR) is able to efficiently exploit the spatio-temporal information of a shortly observed sequence (400ms) and generates a competitive accuracy against the current state-of-the-art. 2CH-TR stands out for the efficient performance of the Transformer, being lighter and faster than its competitors. In addition, our model is tested in conditions where the human motion is severely occluded, demonstrating its robustness in reconstructing and predicting 3D human motion in a highly noisy environment. Our experiment results show that the proposed 2CH-TR outperforms the ST-Transformer, which is another state-of-the-art model based on the Transformer, in terms of reconstruction and prediction under the same conditions of input prefix. Our model reduces in 8.89\% the mean squared error of ST-Transformer in short-term prediction, and 2.57\% in long-term prediction in Human3.6M dataset with 400ms input prefix. Visit our website \href{https://sites.google.com/view/estevevallsmascaro/publications/iros2022}{here}.

\end{abstract}

\section{Introduction}

Humans have the capacity to forecast future states of affairs based on own-constructed models of physical and socio-cultural systems. This ability is developed in childhood through observation and active participation in society. For instance, humans are able to anticipate the movement intention of the others and act accordingly to perform a given task efficiently. This capacity can even work in conditions where the view of the other people is partially occluded.

\begin{figure}
\centering
\includegraphics[width=0.40\textwidth]{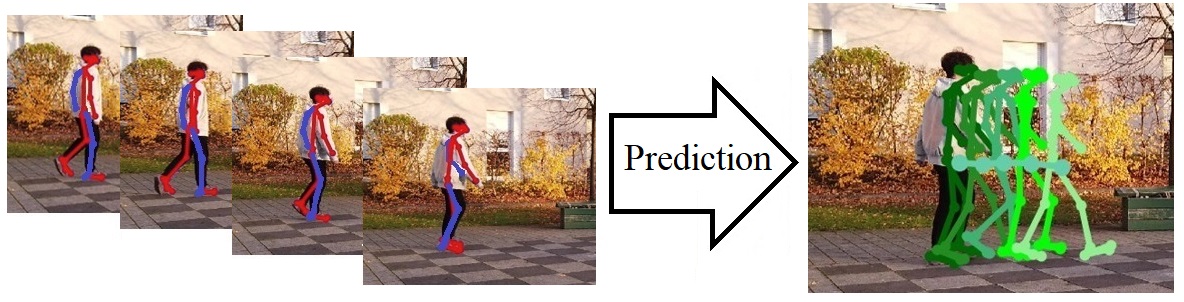}
\caption{An overview of 3D human motion forecasting in occluded environments. The red lines are the observed 3D skeletons projected into the image, while the blue lines consist of random occluded limbs to test model's 3D pose-reconstruction capacity. Finally, the green skeletons represent the predicted human pose sequence in the near future.} \label{fig:intro}

\end{figure}

For robots to coexist with humans, it is also crucial to successfully anticipate nearby human's future movement in real time, even though the view is partially occluded. Then, the robot can adapt its behaviour accordingly to assist the human. Regarding this, 3D Human Motion Forecasting is the research field aimed at predicting the human's future full-body 3D trajectory based on past observations. As shown in Fig.~\ref{fig:intro}, the goal of this research is to generate a possible sequence of future 3D actions based on the short observation of the human body. Then, it would be possible for robots to plan its motion in advance, so that natural co-existence with humans can be realized. To do this, robots need a computationally efficient algorithm that can operate in real-time. However, recently there is a general tendency in the computer vision research community that larger and heavier models are preferred, thus hindering their applicability in robotics. Therefore, our paper proposes a 3D human motion forecasting model that stands out for being faster and lighter than the state-of-the-art, with a similar or even higher performance in the very short term (i.e. 400ms).

3D human motion forecasting is a long-standing challenge that has been addressed by exploiting the spatial-temporal dependencies in the observed skeletons. However, before using spatial and temporal information together, initial works focused only on the influence of temporal history in motion to forecast future human poses. They used recurrent neural networks (RNNs) \cite{RNN, S-RNN} to model dependencies between the skeletons in time and allowed the propagation of information for the short- and long-term human motion forecasting. However, these auto-regressive models accumulated an error over time so that the result eventually collapsed into unrealistic human poses as argued in \cite{autoregressive1, autoregressive2}.

\par

\begin{figure}
\centering
\includegraphics[width=0.40\textwidth]{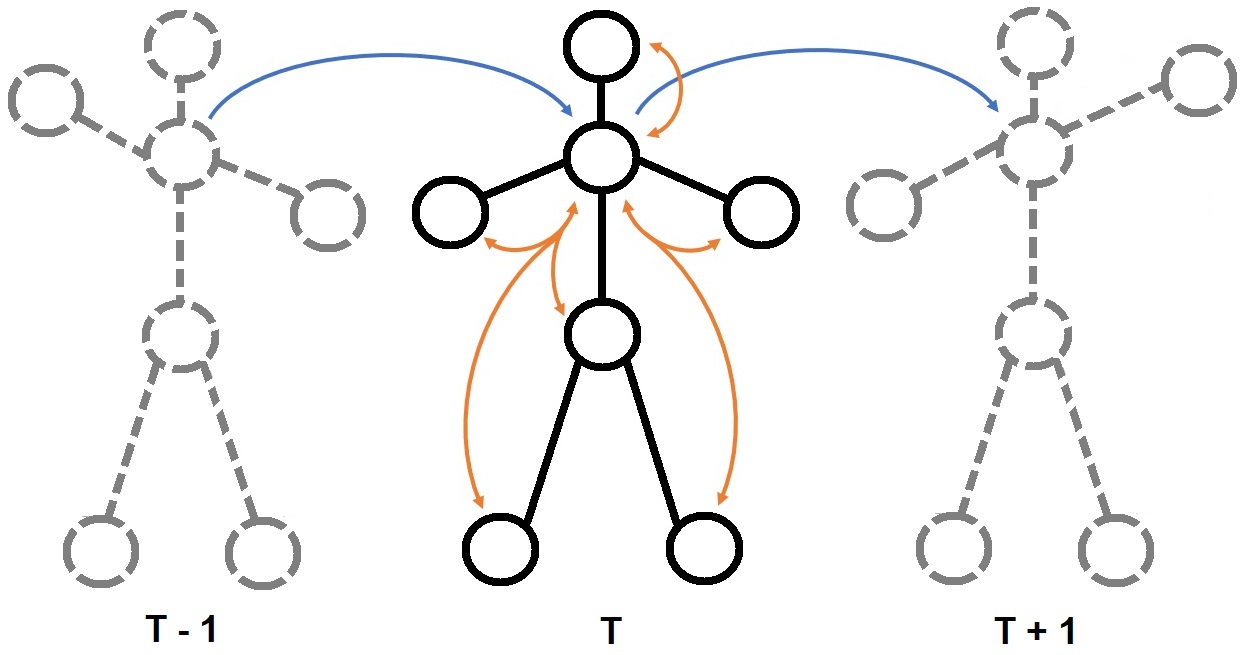}
\caption{Spatio-temporal graph of joint dependencies for human motion. The blue arrows refer to temporal relationships between the same joint parameters in different frames. The orange arrows imply the spatial relationship between joints in the same frame.}
\label{fig:stgraph}
\end{figure}

To avoid forecasting unfeasible poses, it is necessary for models to also understand the spatial dependencies between different parts of the human skeleton. Consequently, several methods \cite{Non-rigid, DCTSim, DCTAttention} attempted to exploit the correlation between different joints or limbs while conserving time dependency. This spatial-temporal approach can be observed in Fig.~\ref{fig:stgraph}, where the dependencies in space (orange) and time (blue) of a given joint are described through a graph. This graph-based approach was proposed as a model architecture to exploit the natural structure of the kinematic tree. For instance,  DCT-GCN \cite{DCT-GCN} encoded the temporal information in feed-forward networks through discrete cosine transformation (DCT) and captured the spatial component of human movement through a learnable graph convolution network (GCN). However, this approach failed in modelling diverse long-sequence as relied on fixed DCT coefficients.  Motivated by the great advances of the Transformer model in language modelling \cite{SBERT, GPT3}, and to improve exploiting long-term dependencies, ST-Transformer \cite{ST-TRANSFORMER} was also proposed to capture this space-time duality of a human body using the self-attention mechanism \cite{ATTENTION}.

However, none of these methods focuses on the application of human motion forecasting in robotics. DCT-GCN \cite{DCT-GCN} dealt with different models for short-term and long-term prediction of 3D joint trajectory, and also lacked in forecasting global rotation of the human while moving. Then, robots could not predict, for instance, towards where the human was walking. ST-Transformer \cite{ST-TRANSFORMER}, in addition, needed a 2-second input sequence to produce a prediction, thus depending on longer observations and increasing the computational resources of the model in terms of size and time. To sum up, none of these existing models is feasible for stand-alone implementation for robotics working in real world.

Moreover, to fairly assess the forecasting capacity of robots and facilitate its incorporation in real environments, the motion forecasting should also be tested in noisy and strongly occluded situations. Robustness to occlusions or noise in observed human skeleton is essential in real application where the 3D human motion forecasting is based on the results obtained from 3D human skeleton estimation. Unlike previous works mentioned above, our work also studies the reconstruction and prediction capacity of models against different types of strong occlusions in the observed input sequence. Our proposed model, named as 2 Channel-Transformer (2CH-TR), allows to cope with high levels of occlusions through its independence in 2 channels, and promotes the 3D motion forecasting, taking into account its applicability in the field of robotics. 

\par
To extensively show the effectiveness of the proposed model, we conduct a quantitative comparison experiment with the state-of-the-art models based on the Human3.6M Dataset \cite{HUMAN3.6m}.  Results show that our 2CH-TR obtains competitive results with state-of-the-art model and outperforms Transformer-based approaches in all time horizons. Under the same prefix input length (400ms), our work reduces ST-Transformer's mean squared error in 8.89\% for short-term forecasting, and 2.57\% for long-term forecasting. In order to test the model performance in real scenario, qualitative results over a real video on our own data set is also reported.

\par

Our contributions can be summarized as follows: (i) we propose a light and fast model for efficient 3D human motion forecasting, suitable for robotic applications; (ii) we construct a model that deals with short- and long-term predictions (from 0 to 1000ms) in a single time (no autorregressive method) while working with simple input prefix pattern (400ms) ; (iii) to consider the noisy environment in the real world, we also study the influence of occlusions and show that the proposed 2CH-TR can adequately estimate the complete human skeleton successfully even with severe occlusions.

\vspace*{-1mm}
\section{Related work}
\begin{figure*}
\centering
\includegraphics[width=1\textwidth]{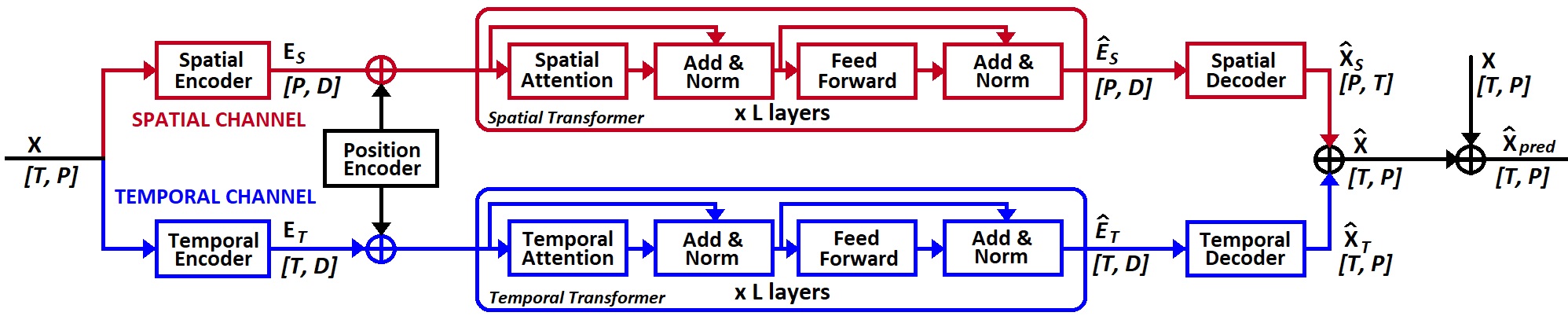}
\caption{Architecture of 2-Channel Transformer (2CH-TR). The observed skeleton motion sequence $X$ is projected independently for each channel into an embedding space ($E_S$ and $E_T$) and then positional encoding is injected. Each embedding is fed into $L$ stacked attention layers that extracts dependencies between the sequence using multi-head attention. Finally, each embedding ($\hat{E}_S$ and $\hat{E}_T$) is decoded and projected back to skeleton sequences. Future poses ($\hat{X}_{pred}$) are then the result of summing the output of each channel ($\hat{X}_S$ and $\hat{X}_T$) with the residual connection $X$ from input to output.}
\label{fig:2ch}
\end{figure*}

\subsection{Recurrent-based approach}

Recurrent neural networks (RNNs) \cite{RNN,S-RNN, RNN3} and long short-term memory (LSTM) \cite{LSTM} were the dominant architectures for modeling temporal dependencies between the human skeletons. Jain et. al. explicitly emphasized the importance of understanding the human body structure to exploit relationships between limbs more effectively through Structured-RNN (S-RNN) \cite{S-RNN}. This work attempted to better explore inter- and intra-relationships between each part of the skeleton, focusing on spine-arms and spine-legs correlation.

Since these recurrent models encoded the history of motion in a hidden state of fixed size, an error could be propagated through time and long-term dependencies could not be exploited efficiently. To solve these problems, various works applied data augmentation tricks through Gaussian noise to the inputs \cite{RNN3, longterm} or used adversarial losses \cite{autoregressive1, autoregressive2} to tackle this vanishing of information, but their long-term predictions still collapsed into non-plausible poses. Compared to this, our work takes advantage of self-attention mechanisms, that avoid compressing all historical information into a fixed-length hidden state, and have an ability to attend every historical pose at every time step. Therefore, our predictions are made by weighting which states are more informative at every step in the sequence, excelling in exploiting long-term dependencies and tackling the vanishing gradient challenge of recurrent networks.

\par
\subsection{Spatio-Temporal Modeling}

State-of-the-art approaches tackled 3D human motion forecasting by leveraging spatio-temporal dependencies. Then, graph structures were proposed as spatial representations of human’s body to exploit the natural structure of the kinematic tree \cite{GraphST}. However, they failed when capturing human motions that required synchronization between limbs, such as the periodic movement between arms and legs when walking. Consequently, these non-apparent dependencies in kinematic-tree structures needed to be learned based on the data. To tackle that, graph convolutional networks (GCN) emerged to adaptively capture these connections necessary for motion forecasting \cite{DCTAttention, DCT-GCN}. State-of-the-art DCT-GCN \cite{DCT-GCN} model leveraged GCN to encode the joint relationships, and adopted discrete cosine transforms (DCT) to capture the smoothness of motion in time. However, their work only attempted to predict joint motion, failing in forecasting global rotation of the humans, and therefore not being appropriate for its application in the robotics field. Moreover, \cite{DCT-GCN} trained different models for each time horizon (short and long-term predictions), thus increasing the required computational resources for predicting in both scenarios.

Inspired by recent advances of self-attention mechanisms in natural language processing (NLP) \cite{SBERT, GPT3}, Emre et. al. introduced the Transformer-based architecture into human motion forecasting and emphasized the concepts of time and space by designing a spatio-temporal Transformer (ST-Transformer) \cite{ST-TRANSFORMER}. Their aim was to take advantage of the low inductive bias of the transformer-based architectures shown in language modelling and nowadays also in computer vision \cite{InductiveBias}. The success of this architecture had two main reasons. First, transformers exploited capturing both short and long term dependencies by using the positional encoding \cite{ATTENTION}.  Second, by adding Multi-Head Attention \cite{ATTENTION}, the model was able to extract richer dependencies from the observed sequence by  attending in parallel to different representation sub-spaces.

Our proposed model is inspired by the efficiency of ST-Transformer but differs in essence. For example, while ST-Transformer deals with joint-wise vectors formed by a rotation matrix representation as inputs, and over-emphasized relationships between these joints, we rather propose a model that autonomously learns these relationships from all flatted skeleton parameters, without previous clustering into joints. Our \textbf{temporal channel} explores these relationships in each time frame, while the \textbf{spatial channel} identifies intra-framed relationships of the skeleton. Total decoupling of our two channels allows to boost the robustness of the model while simplifying the structure. Moreover, our work is also different in terms of input pattern. ST-Transformer requires 2-seconds human motion observations (50 frames at 25 FPS) as an input to produce 1-second prediction in Human3.6M dataset, while ours reduce it by to 20\% the needed input length (400ms, 10 frames at 25 FPS) for the same time-length prediction.

\subsection{Skeleton Recovery}
Application of stand-alone state-of-the-art 3D human motion forecasting in a real environment might be unfeasible if there are numerous occlusions or noise in observed skeleton data. Existing works do not mention about the capacity of their models to work under occluded environments. 

However, this topic is indeed researched in the field of pose estimation, by recovering the occlusion when estimating the body pose. For instance, Guo et. al. employed a fully convolutional network (FCN) to enable the regression between occluded and complete distance matrix~\cite{recovery} for pose estimation. Then, Cao et. al. with OpenPose~\cite{OPENPOSE} used Part Affinity Fields (PAFs) to perform skeleton estimation in the presence of human occlusion, but this was limited to 2D poses. Cheng Yu et. al. worked with occlusion-aware convolutional neural networks, named Cylinder Man Model \cite{CYILINDERMAN}, to mitigate the effect of occlusions. All of these models dealt with occlusion and noise recovery in the field of skeleton estimation, but none in the 3D human motion forecasting.

Finally, Ruiz et. al. designed a model based on Generative Adversarial Networks (GAN), named as MotionGAN~\cite{MOTION-GAN}, and formulate 3D motion forecasting as an in-painting problem by totally masking the future frames to predict. Then, MotionGAN attempted both human motion reconstruction and forecasting independently, but never predict the future motion based on a occluded observation. Our work evaluates the capacity of 2CH-TR to forecast the future, also when the observed input is partially occluded,

To the best of our knowledge, our work proposes a new line of research based on the evaluation and improvement of 3D human motion forecasting in highly occluded environments. We claim the importance of our investigation to facilitate the future implementation in robotic-oriented scenarios.
\vspace*{-1mm}
\section{Method}

Human motion forecasting needs to exploit spatio-temporal dependencies to generate plausible future poses. The encoder-decoder structure inherited from Transformers is proposed in our 2CH-TR, inspired by \cite{ST-TRANSFORMER}. Fig. \ref{fig:2ch} shows the overview of the proposed architecture, that clearly represents the independence between our spatial channel (in pink lines) and temporal channel (in light blue lines).

\par

\subsection{Problem Formulation}
\begin{figure}
\centering
\includegraphics[width=0.45\textwidth]{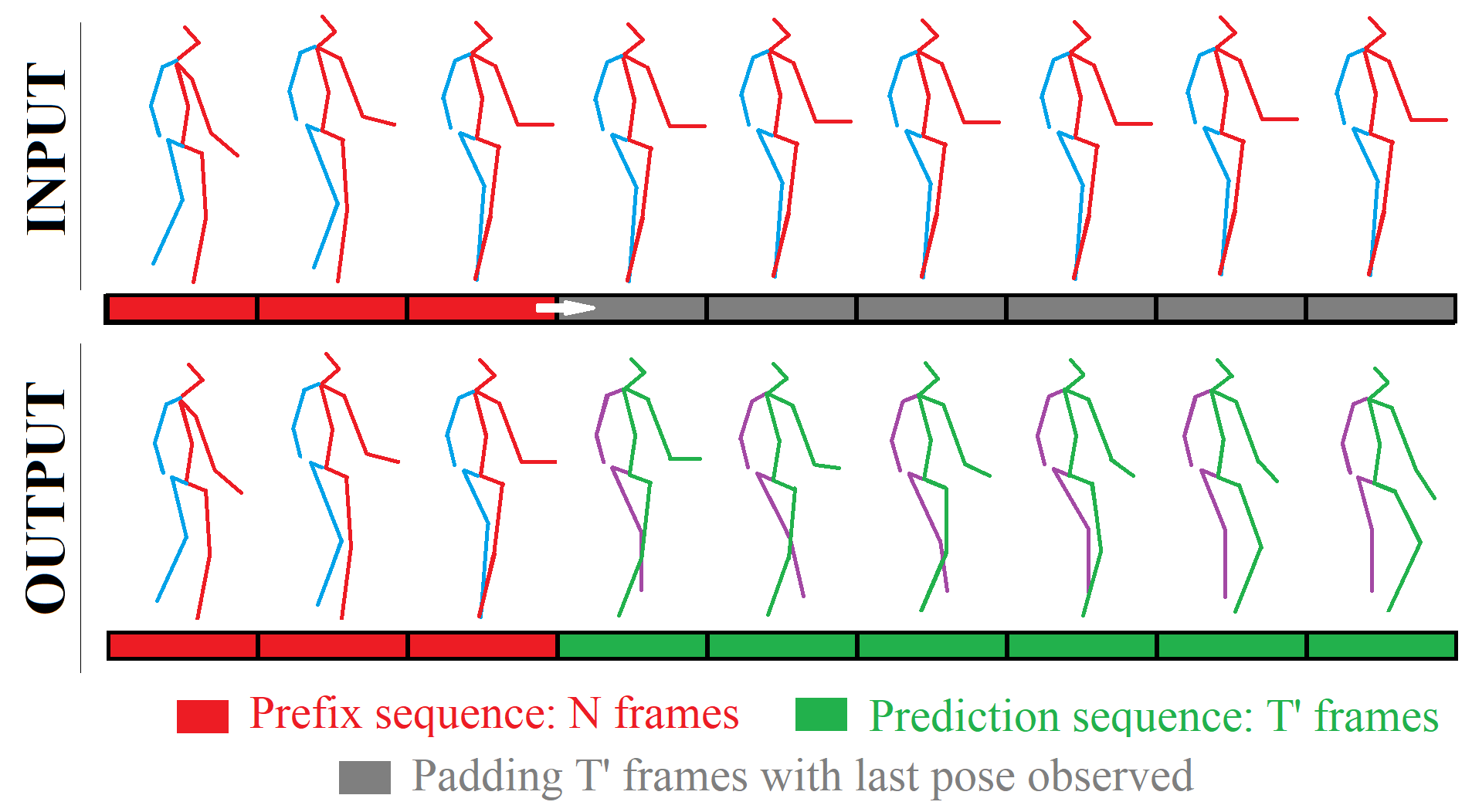}
\caption{Input pattern representation for our 2CH-TR, with $N=10$ poses observed by the model and $T'=25$ future poses to be predicted. In the prefix, last $T'$ poses are repeated from last observed pose, so that the estimation only focuses on forecasting the difference between the future pose and final pose.}
\label{fig:pattern}
\end{figure}

Let $x_t=[x_{t,1},\cdots,x_{t,P}] \in \mathrm{R}^{P}$ denote our skeleton parameter at time frame $t$, that defines a set of human joints in axis-angle representations. Here, $x_{t,1},\cdots,x_{t,3}$ consist on the global rotation information of the human body, so that our model can also learn how to handle orientation. Given an observed motion sequence $X_{1:N}$, named as prefix, we replicate last pose $x_N$ for $T'$ times to generate a prediction of future $T'$ frames, obtaining a final sequence $T = N +T'$, as it is described in Fig. \ref{fig:pattern}. This padding pattern diminishes the complexity of the model as it only needs to learn the variation of each skeleton parameter based on the lastly observed pose.

Then, the final motion sequence is represented as a matrix $X = [x_1,\cdots,x_T] \in \mathbf{R}^{T\times P}$,  where $T$ represents the number of time frames in the sequence and $P$ indicates the number of skeleton parameters. Unlike \cite{ST-TRANSFORMER}, which classifies and over-emphasizes relationships among different joints, our 2CH-TR exploits all temporal and spatial dependencies independently as a whole, projecting the prefix sequence into an independent embedding for each of the 2 channels ($E_S$ and $E_T$) and only coupling them for the final result. Our motivation is to decompose the contribution of each channel until the last stage to give robustness and improve short- and long-term predictions. Compared to \cite{ST-TRANSFORMER}, our approach provides higher performance with lower number of attention layers $L$ and using only $20$\% of input prefix $N$, which also reduces the dimension of the model.  Moreover, our model takes advantage of the encoder-decoder structure to generate the whole sequence in a single prediction, avoiding frame-by-frame prediction as in \cite{ST-TRANSFORMER}, thus reducing inference time.

\subsection{Temporal Channel}
\begin{figure}
\centering
\includegraphics[width=0.47\textwidth]{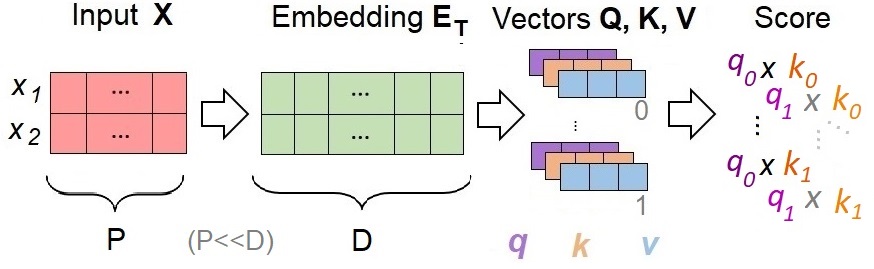}
\caption{Temporal channel mechanism to exploit relationships of $P$ skeleton parameters between $N$ frames. Attention is used to capture time dependencies in the projected embedding space. For simplification in the visualization, only a historical of $T=2$ poses ($x_1$ and $x_2$) are used.}
\label{fig:tch_ex}
\end{figure}
\par

As shown in Figure~\ref{fig:tch_ex}, the skeleton parameters used to describe the human pose are projected from the $P$ dimension to the $D$ dimension $(D>>P)$ via the temporal encoder to learn the context of the timeline for each dimension. Then, the embedding $E_T\in\mathbf{R}^{T\times D}$ is passed into the temporal transformer after being ordered by sinusoidal positional encoding. Multi-head attention (MHA) used in this Transformer block jointly leverages the relationship of the same joint in all its time-history. Each attention head $head_{T}^{(h)}$, where $h \in(1,\cdots ,H)$, linearly projects the query $(Q)$, key $(K)$ and value $(V)$ obtained from embedding $E_T$ by employing three different learnable weight matrices as shown in equation ~\eqref{eq:2c_t_1}, where $Q_{T}^{(h)}, K_{T}^{(h)}, V_{T}^{(h)} \in \mathbf{R}^{T\times F}, F=D/H$, refer to the query, key and value matrix of $head_{T}^{(h)}$ in temporal transformer block, respectively.  $W_{T}^{(Q,h)},W_{T}^{(K,h)},W_{T}^{(V,h)}\in \mathbf{R}^{D\times F}$ indicate the learnable weight matrices for $head_{T}^{(h)}$.

\vspace*{-4mm}
\begin{equation}
\begin{split}
& Q_{T}^{(h)}= E_{T}W_{T}^{(Q,h)} \\ 
& K_{T}^{(h)}= E_{T}W_{T}^{(K,h)} \ ,\ h\in(1,\cdots ,H) \\
& V_{T}^{(h)}= E_{T}W_{T}^{(V,h)}\\
\end{split}
\ , \label{eq:2c_t_1}
\end{equation}
\par
For each head of the Transformer block, the applied attention mechanism is shown in equation~\eqref{eq:2c_t_2}. The relationship between any two time points can be discovered in the attention matrix $A_T\in\mathbf{R}^{T\times T}$. Besides, a temporal mask $M\in\mathbf{R}^{T\times T}$ is applied to guarantee that future information cannot be leaked to the past.

\vspace*{-4mm}
\begin{equation}
\begin{split}
head_{T}^{(h)}& = \textrm{softmax}(\cfrac{Q_{T}^{(h)}K_{T}^{(h)^T}}{\sqrt{D}}+M)V=A_{T}^{(h)}V_{T}^{(h)}
\end{split}
\ , \label{eq:2c_t_2}
\end{equation}

The prediction is projected by concatenating of all heads, as shown in equation~\eqref{eq:2c_t_3}, where $ W_{T}^{(O)}\in\mathbf{R}^{HF\times D}$ denotes the MHA learnable matrix. By $L$ temporal-attention block stacking, the prediction performance can be strengthened.

\vspace*{-3mm}
\begin{equation}
\hat{E}_{T}= [head_{T}^{(1)},\cdots  , head_{T}^{(H)}] W_{T}^{(O)}\ , \label{eq:2c_t_3}
\end{equation}

\par

Fig.~\ref{fig:tch_ex} explains how our temporal channel works with a simplified example, where the input sequence is assumed of only 2 time frames ($X\in\mathbf{R}^{2\times P}$). In temporal channel, we mainly focus on the relationship between each frame, i.e, the relationship between $x_{1}$ and $x_{2}$. After the temporal encoder, the temporal embedding $E_{T}\in\mathbf{R}^{2\times D}$ multiplies with three different weight matrices to obtain query, key, and value matrices. Each row of these matrix represents one frame. When $x_{1}$ is presented as a query, the self-attention compares all frames in the sequence as keys to $x_{1}$ and generates a correlation or attention score. A larger score represents a higher degree of correlation. Then, this score is normalized and multiplied with the value vector to get the final output, which encodes the importance of the dependency. To this end, temporal channel can learn the joint relationships between each time-frame.

\subsection{Spatial Channel}
\begin{figure}
    \centering
    \includegraphics[width=0.47\textwidth]{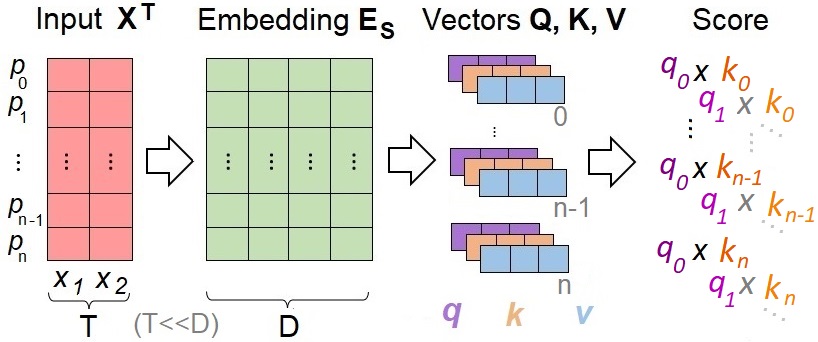}
    \caption{Spatial Channel projects $T$ dimension to $D$ embeddings to focus on the relationships between the parameters at each time horizon. Then, attention is applied to capture dependencies of skeleton parts at each pose. For simplification, only $T=2$ historical poses ($x_1$ and $x_2$) are used.}
    \label{fig:sch_ex}
\end{figure}

While the temporal channel efficiently captures the evolution of parameters in time, a spatial channel is able to understand the underlying dependencies among the skeleton parameters at each time horizon, to represent the plausible poses of a human. 
For the spatial channel, as described in Fig.~\ref{fig:sch_ex},  the skeleton parameter size $P$ is preserved and the sequence length $T$ is projected into $D$ dimension, where $X \xrightarrow{} E_{S}: \mathbf{R}^{P\times T} \xrightarrow{} \mathbf{R}^{P \times D}$. Similar to the temporal Transformer, the spatial Transformer also has multi-head attention and 3 different learnable weight matrices are utilized to linearly project query, key and value in each head. Since its purpose is to explore spatial relationships, the mask $M$ is no needed in this case. When applying attention, the spatial attention matrix $A_{S}\in \mathbf{R}^{P\times P}$ interprets the relationship of every parameter of the entire skeleton, thus learns the spatial structure of the skeleton under different movements. Through the stack of $L$ spatial-attention blocks, the model can predict the pose under each frame more accurately. Finally, the output embedded representation $\hat{E}_S$ is fed into the spatial decoder and permuted, obtaining the result $\hat{X}_{S}$. The whole process is similar to the Temporal Channel ($T$) expressed by equation~\eqref{eq:2c_t_1}, ~\eqref{eq:2c_t_2}, ~\eqref{eq:2c_t_3}, but using the $S$ notation for the Spatial Channel. The resulting equations show the use of the spatial weight matrices $W_{S}^{(Q,h)}, W_{S}^{(K,h)}, W_{S}^{(V,h)}, W_{S}^{(O)}$, with same dimensions as in temporal channel, and $Q_{S}^{(h)}, K_{S}^{(h)}, V_{S}^{(h)} \in \mathbf{R}^{P\times F}, F=D/H$ encoded now in the space.


Finally, results from two channels' result $\hat{X}_{T}$ and $\hat{X}_{S}$ are summed at the end to obtain the motion prediction $\hat{X}$. Afterwards, the input and output of the model are connected by a residual connection to diminish the possibility of gradient explosion when the model depth is too large.

\vspace{-2mm}
\section{Experiments}
\vspace{-2mm}
\subsection{Datasets}
\textbf{Human3.6M.} H3.6M \cite{HUMAN3.6m} includes 3.6 million 3D human poses from high-resolution videos of 7 subjects performing 15 different actions such as walking, eating and smoking. Body skeleton of each subject is represented by $P=99$ skeleton parameters which include the global rotation and translation of the motion. Global rotation is crucial for robots to understand the intention of humans when performing an action. For a fair comparison with previous works \cite{S-RNN, DCT-GCN, ST-TRANSFORMER}, videos are down-sampled to 25 frames per second and the test is performed on the same sequence of the subject 5.
\par
\textbf{Testing Scenario.} More realistic human motion scenarios are used for demonstrating the effectiveness of our model in the qualitative results. Observation of prefix poses are obtained through FrankMocap \cite{FRANKMOCAP}, thus realizing real-world application.

\subsection{Evaluation Metrics and Baselines}
\textbf{Metrics.} Following the standard evaluation protocols, we report the mean squared error (MSE) metric, as shown in equation~\eqref{eq:error}, between all the predicted $\hat{p}^{i}$ and ground-truth $\hat{p}^{i}$  joint angles in Euler angle representation, where  $i \in(1,\cdots ,N)$, and $N$ is the number of joints used to describe the human poses.

\vspace*{-5mm}
\begin{equation}
    error(X_{targ}, \hat{X}_{pred}) = \sum_{i=1}^N \lvert\lvert p^{i}-\hat{p}^{i} \rvert\rvert ^2
\label{eq:error}
\end{equation}
\vspace*{-3mm}

\textbf{Baselines.} We compare 2CH-TR with state-of-the-art methods S-RNN \cite{S-RNN}, DCT-GCN \cite{DCT-GCN} and ST-Transformer \cite{ST-TRANSFORMER} in Human3.6M dataset. We implement their work with original code provided by the authors and evaluate the models under the same condition for fair comparison. Details of how existing models are trained are as follows. DCT-GCN was trained independently for a short- and long-term joint prediction, while we propose a stand-alone model that deals simultaneously with both time horizons to reduce computational resources and to fasten the inference. Moreover, 2CH-TR also predicts global rotation of the human, apart from joint trajectories, to capture the orientation of real motion. ST-Transformer needs the input prefix sequence of 2 seconds for 1 second prediction. Compared to this, we manage the same-length prediction only with 400ms prefix sequence, so that the complexity of our approach can be reduced by relying on less information (and also reduces the dimension of the architecture, being more efficient).

\subsection{Skeleton Occlusion and Recovery}
\label{chap:occlusions}
\begin{figure}
    \centering
    \includegraphics[width=0.45\textwidth]{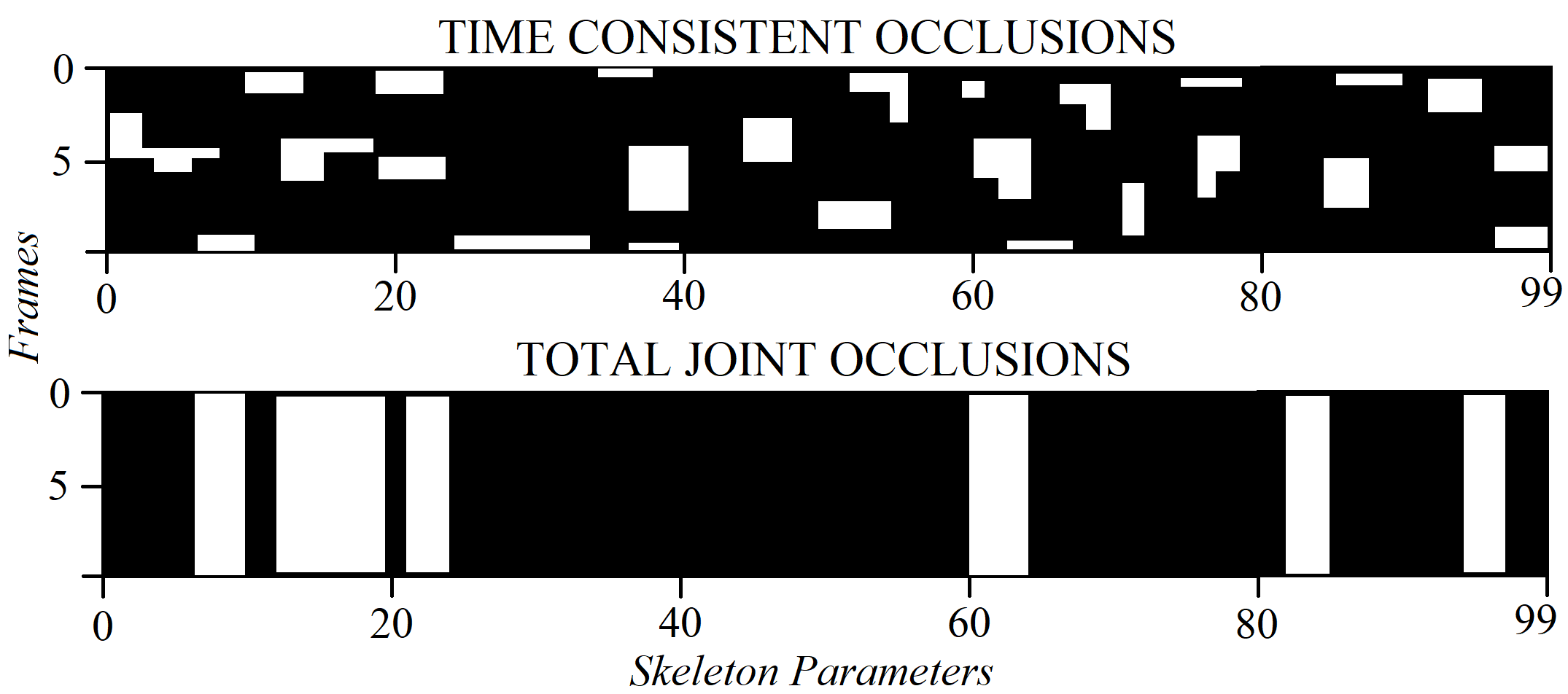}
    \caption{Visualization of randomized occlusions for observed prefix motion. In this example, 80\% of data is missing (black colour denotes occluded data).}
    \label{fig:occl}
\end{figure}
The motivation behind this paper consists of building a model applicable in real-world robotics. To our knowledge, the feasibility of this task consists of evaluating four important concerns: (i) capacity of the model to perform short- and long-term motion forecasting in ideal scenarios; (ii) performance of the model when tested in noisy scenarios (i.e. occlusions in prefix sequences); (iii) efficiency of the algorithm regarding the inference time; (iv) lightness of the model for usage in a real robot without compromising its hardware capacities.

To reproduce scenarios of noisy prefix observations, we propose two different types of occlusions that will lead our following evaluation. First, as shown in upper-side of Fig. \ref{fig:occl} we defined Time Consistent Occlusions to reproduce randomly partial missing joints (visualized as black) in the prefix sequence, similar to possible errors when estimating human skeleton. Each occlusion has a time duration based on an exponential distribution. The second scenario is the total occlusion of several random joints in the whole prefix observed motion, as illustrated in the lower-side of Fig. \ref{fig:occl} where random skeleton parameters are missing (visualized as black), simulating real-life occlusions, where a part of the body, for instance, cannot be observed as being covered by the environment. Fig. \ref{fig:occl} exemplifies randomized occlusions in our $N=10$ frames prefix motion (400ms) for all the $P=99$ skeleton parameters provided by Human3.6M dataset.

\par

We evaluate our model when the percentages of occluded joints vary among 80\%, 60\%, 40\% and 20\%. To reconstruct occlusion in observed motion  (of the past -400ms to present, 0 ms), several mechanisms are tested: (i) short-term recovery which predicts occluded information from previous non-occluded motion (based on assumed already reconstructed sequence in the past, from -800ms to -400ms); (ii) auto-regressive recovery that continuously predicts immediate (+80ms) skeleton pose and refines it with the known non-occluded data in the observation; (iii) linear interpolation applied in time that deals with partial skeleton occlusions to explode bidirectional flow information inherited by the interpolation technique.

\begin{table*}
\centering
\caption{MSE error comparison of human motion forecasting in Human 3.6M dataset. }
\resizebox{\textwidth}{!}{%
\begin{tabular}{llccccc|cccccc|cccccc}
\multicolumn{1}{c}{\textbf{}} &  & \multicolumn{5}{c|}{\textbf{WALKING}}                                                    & \multicolumn{6}{c|}{\textbf{PHONING}}                                                                                                   & \multicolumn{6}{c}{\textbf{WAITING}}                                                                 \\
milliseconds (ms)             &  & 80             & 160            & 320            & 560            & 1000                 & 80                   & 160                  & 320                  & 400                  & 560                  & 1000                 & 80             & 160            & 320            & 400            & 560            & 1000            \\ 
\hline
\textbf{S-RNN}                &  & 0.808          & 0.942          & 1.159          & 1.484          & 1.778                & 1.225                & 1.503                & 1.925                & 2.061                & 2.02                 & 2.38                 & 1.156          & 1.396          & 1.781          & 1.941          & 2.191          & 2.957           \\
\textbf{ST-Transformer}       &  & 0.212          & 0.359          & 0.58           & 0.72           & 0.782                & 0.53                 & 1.042                & 1.41                 & 1.544                & 1.543                & 1.809                & \textbf{0.219} & 0.512          & 0.978          & 1.221          & 1.658          & 2.485           \\
\textbf{DCT-GCN}              &  & \textbf{0.201} & \textbf{0.344} & \textbf{0.516} & \textbf{0.647} & \textbf{0.673}       & 0.541                & 1.026                & 1.342                & 1.472                & 1.454                & \textbf{1.649}       & 0.252          & 0.517          & 0.957          & \textbf{1.169} & \textbf{1.546} & \textbf{2.293}  \\ 
\hline
\textbf{2CH-TR (ours)}        &  & 0.204          & 0.357          & 0.57           & 0.745          & 0.908                & \textbf{0.526}       & \textbf{0.982}       & \textbf{1.238}       & \textbf{1.373}       & \textbf{1.406}       & 1.725                & 0.237 & \textbf{0.508} & \textbf{0.936} & 1.17           & 1.607          & 2.312           \\
\textbf{}                     &  &                &                &                &                & \multicolumn{1}{c}{} & \multicolumn{1}{l}{} & \multicolumn{1}{l}{} & \multicolumn{1}{l}{} & \multicolumn{1}{l}{} & \multicolumn{1}{l}{} & \multicolumn{1}{l}{} &                &                &                &                &                &                 \\
\multicolumn{1}{c}{\textbf{}} &  & \multicolumn{5}{c|}{\textbf{WALKING DOG}}                                                & \multicolumn{6}{c|}{\textbf{POSING}}                                                                                                    & \multicolumn{6}{c}{\textbf{PURHCASES}}                                                               \\
milliseconds (ms)             &  & 80             & 160            & 320            & 560            & 1000                 & 80                   & 160                  & 320                  & 400                  & 560                  & 1000                 & 80             & 160            & 320            & 400            & 560            & 1000            \\ 
\hline
\textbf{S-RNN}                &  & 1.029          & 1.221          & 1.546          & 2.067          & 2.471                & 1.346                & 1.395                & 1.96                 & 2.2                  & 2.456                & 3.102                & 1.219          & 1.452          & 1.87           & 1.989          & 2.36           & 3.325           \\
\textbf{ST-Transformer}       &  & \textbf{0.43}  & \textbf{0.783} & 1.148          & 1.613          & 1.896                & 0.609                & 0.684                & 1.052                & 1.282                & 1.776                & 2.826                & \textbf{0.43}  & 0.765          & 1.304          & 1.373          & 1.548          & 2.411           \\
\textbf{DCT-GCN}              &  & 0.489          & 0.804          & \textbf{1.11}  & 1.525          & \textbf{1.841}       & \textbf{0.212}       & \textbf{0.47}        & 1.071                & 1.306                & 1.617                & \textbf{2.42}        & 0.497          & 0.718          & \textbf{1.062} & \textbf{1.121} & \textbf{1.415} & 2.215           \\ 
\hline
\textbf{2CH-TR (ours)}        &  & 0.476          & 0.821          & 1.139          & \textbf{1.524} & 1.916                & 0.226                & 0.51                 & \textbf{1.014}       & \textbf{1.244}       & \textbf{1.595}       & 2.514                & 0.441          & \textbf{0.698} & 1.161          & 1.201          & 1.493          & \textbf{2.196} 
\end{tabular}
}
\label{tab:results}
\end{table*}
\subsection{Results}
\textbf{Quantitative Evaluation}

We report our results for short-term ($<$500ms) and long-term ($>$500ms) predictions in Human3.6M dataset. We maintain the observation prefix patterns same as each baseline originally used. Also, the MSE error is measured over all parameters predicted by each model. Note that our model also includes the error of global rotation prediction.

In Table \ref{tab:results}, we compare our model with existing works in terms of several actions consisting the Human3.6M dataset. For example, actions such as `Walking', `Phoning', `Waiting', `Walking Dog',  `Posing', and `Purchases' are considered. The results show that our model obtains better or very competitive results in short-term prediction, outperforming ST-Transformer in most cases.  The MSE report in Table \ref{tab:average} shows the competitive performance of our model when predicting human motion in non-occluded environments.


%

\begin{table}
\caption{Average MSE error for subject S5 in motion prediction}

\resizebox{0.50\textwidth}{!}{%
\begin{tabular}{lcccccc}
milliseconds (ms)\textbf{}                & \textbf{80}    & \textbf{160}   & \textbf{320}   & \textbf{400}   & \textbf{560}   & \textbf{1000}   \\ 
\hline
\textbf{S-RNN (orig.)}           & 0.933          & 1.166          & 1.397          & 1.526          & 1.711          & 2.139           \\
\textbf{S-RNN (N=10)}          & 0.988          & 1.161          & 1.435          & 1.576          & 1.84           & 2.221           \\
\textbf{DCT-GCN}                    & \uline{0.295}  & \textbf{0.542} & \textbf{0.857} & \textbf{0.974} & \textbf{1.154} & \textbf{1.590}  \\
\textbf{ST-Transformer (orig.)}  & 0.303          & \uline{0.550}  & 0.901          & 1.021          & \uline{1.229}  & \uline{1.722}   \\
\textbf{ST-Transformer (N=10)} & 0.341          & 0.619          & 0.966          & 1.100          & 1.314          & 1.754           \\ 
\hline
\textbf{2CH-TR (ours)}              & \textbf{0.293} & 0.555          & \uline{0.893}  & \uline{1.016}  & 1.245          & 1.744          
\end{tabular}
}
\label{tab:average}
\end{table}


 As detailed in Sec. \ref{chap:occlusions}, our model is also evaluated in occluded scenarios. Table \ref{tab:partial_occlusion} shows results with Time Consistent occlusions, and it indicates the capacity of our model to overcome errors in human pose estimations in real scenarios (i.e. joints not detected). Our model is tested with 3 different reconstruction methods for occlusion recovery. Results show the robustness of 2CH-TR for human motion forecasting in occluded scenarios. Linear interpolation in time also results as a robust method in our recovery scenario as the variation of joint movement in short occluded periods is not significant. However, linear interpolation uses bidirectional information (skeletons before and after the occlusion) to recover the joint position, in contrast to the undirectionality of our 2CH-TR. This behaviour is desired for real-world applications, where the model cannot rely on future data as we do not have it. Auto-regressive approaches for occlusion reconstruction using our 2CH-TR collapse in non-plausible poses in prediction. As our model only forecasts the difference between the future poses and the last pose observed (replicated in the input prefix), the correct reconstruction of this last pose is essential, as any error in it is propagated to the next prediction.


\begin{table}
\caption{MSE error prediction with Time-Consistent  80\% Occlusion }

\resizebox{0.50\textwidth}{!}{%
\begin{tabular}{lcccccc}
milliseconds (ms)                   & \textbf{80}    & \textbf{160}   & \textbf{320}   & \textbf{400}   & \textbf{560}   & \textbf{1000}   \\ 
\hline
\textbf{Model Prediction}           & 1.126          & 1.234          & \textbf{1.386} & \textbf{1.463} & \textbf{1.613} & \textbf{1.960}  \\
\textbf{Linear Interpolation}       & \textbf{0.951} & \textbf{1.177} & 1.443          & 1.526          & 1.650          & 2.015           \\
\textbf{Auto-Regressive Prediction} & 4.039          & 3.738          & 3.637          & 3.579          & 3.495          & 3.424          
\end{tabular}
}
\label{tab:partial_occlusion}
\end{table}

\begin{figure}
    \centering
    \includegraphics[width=0.47\textwidth]{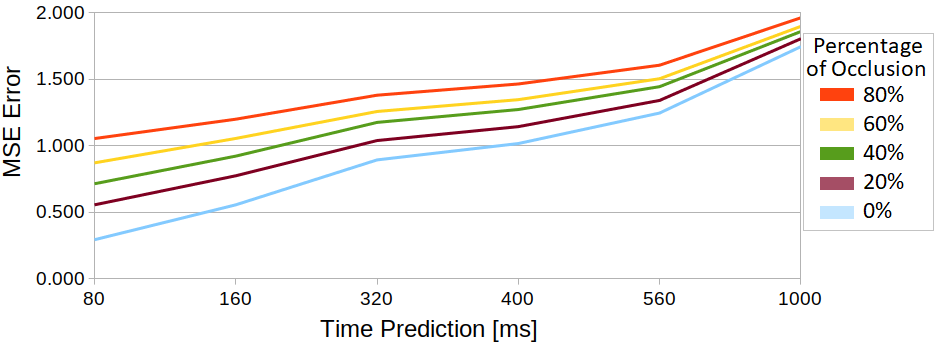}
    \caption{Human Motion Prediction in different occluded environments, from non-occluded data to whole prefix sequence occluded.}
    \label{fig:occludedresults}
\end{figure}

Despite the competitive results of linear interpolation when joints are shortly occluded, this approach requires non-occluded data and cannot reconstruct poses when a skeleton parameter is occluded during the whole observed sequence. Therefore, to assess this second type of occlusion, in Fig. \ref{fig:occludedresults} only the effect in human motion forecasting of reconstructing the missing joints in the whole observed sequence with 2CH-TR is shown. 2CH-TR is able to predict plausible human poses and reliable results even though we test the model in heavily occluded situations, with 80\% or more of joint data occluded in the prefix sequence.

\par

Finally, we assess the contribution of our model in comparison with baseline models regarding the applicability in real-scenario. Computation speed and lightness are reported in Table \ref{tab:applicability}. For a fair evaluation, we compared the size of the baseline networks and their speeds with their original codes but also with reduced prefix pattern to 10 frames as same as ours. The results show our 2CH-TR is the fastest model by increasing the speed over DCT-GCN by 30\% with a GPU Tesla-K80 and around $128$ times faster than ST-Transformer, as our model predicts the whole future sequence in a single time, avoiding autoregression. DCT-GCN is lighter when predicting only short or long-term sequences, but it needs one model for each case, resulting in double-size network in total. As 2CH-TR explicitly performs short- and long-term prediction in one-shot, our model ends up being the lightest in our comparison (59\% lighter that DCT-GCN).

\begin{table}
\centering
\caption{
        Evaluation of model efficiency for 1 second prediction}
\label{tab:applicability}
\resizebox{0.50\textwidth}{!}{%

\begin{tabular}{lccccc}
                           & \begin{tabular}[c]{@{}c@{}}\textbf{Input}\\\textbf{Size}\end{tabular} & \begin{tabular}[c]{@{}c@{}}\textbf{Global}\\\textbf{~Rotation}\end{tabular} & \textbf{Autoregression} & \begin{tabular}[c]{@{}c@{}}\textbf{Inference}\\\textbf{time~[ms]}\end{tabular} & \begin{tabular}[c]{@{}c@{}}\textbf{Network}\\\textbf{~Parameters}\end{tabular}  \\ 
\hline

DCT-GCN                    & 10                                                                    & \xmark                                                                           & \xmark                       & 2 * 3.20                                                                       & 2 * 2,3M                                                                        \\
ST-Transformer             & 50                                                                    & \xmark                                                                           & \cmark                       & 344.66                                                                         & 3,2M                                                                            \\
S-RNN                      & 50                                                                    & \xmark                                                                           & \cmark                       & 117.80                                                                         & 22,8M                                                                           \\
ST-Transformer & 10                                                                    & \xmark                                                                           & \cmark                       & 283.81                                                                         & 3,2M                                                                            \\
S-RNN          & 10                                                                    & \xmark                                                                           & \cmark                       & 45.34                                                                          & 22,8M                                                                           \\ 
\hline
2CH-TR (ours)              & 10                                                                    & \cmark                                                                           & \xmark                       & \textbf{2.21}                                                                  & \textbf{2,6M}                                                                  
\end{tabular}
}

\end{table}

\begin{figure}
    \centering
    \includegraphics[width=0.40\textwidth]{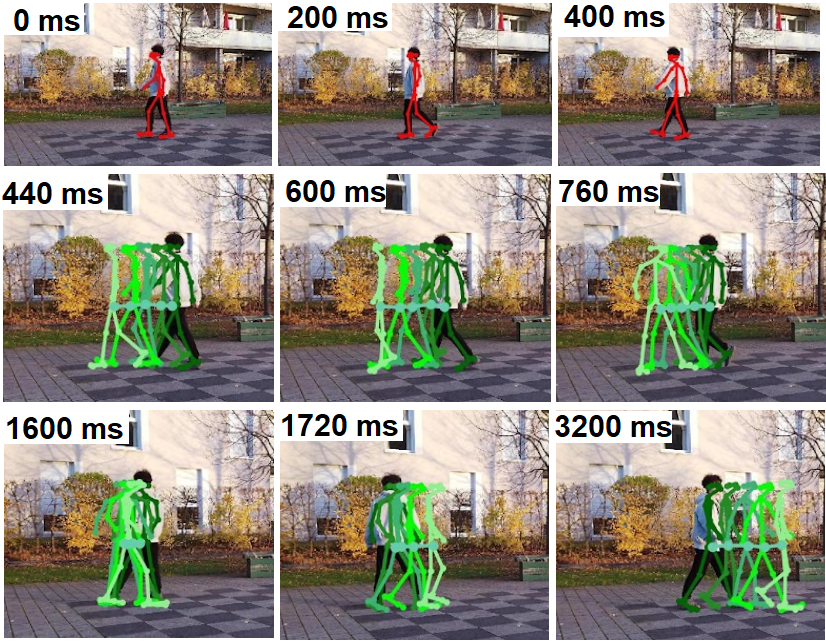}
    \caption{Motion forecasting in the wild when human walks towards the left side, turns around and walks backwards. Red skeleton shows the observed prefix sequence from our model, while gradient green skeletons project human motion prediction in next 1 second.}
    \label{fig:turn_around}
\end{figure}

\textbf{Real world demonstration.}
We tested the performance of our 2CH-TR when forecasting human motion in the wild. For this approach, we trained an additional model with also a global translation parameter (obtained as the center of the 3D skeleton estimated from FrankMocap, located approximately in the torso), so that 2CH-TR can understand the global trajectory of the human motion. Fig. \ref{fig:turn_around}  shows the effectiveness of not only predicting 3D human poses but also global rotation and translation parameters, in contrast with baseline methods. It is shown that the motion of the human is reasonably predicted: 2CH-TR is able to understand the reduction of velocity when the human turns around, and adapt the trajectory and future poses when walks backwards to the new next goal. 

\section{CONCLUSION}

In this work, we propose 2CH-TR architecture to efficiently exploit dependencies between 3D human poses in space and time to forecast near future skeleton sequences. By decoupling the spatial and temporal channels, it is able to tackle high variant action motions in a single prediction. 2CH-TR forecasts 1-second sequence of future poses while only using 400ms of past observations. Our approach obtains competitive state-of-the-art results while reducing the required computational resources and increasing the speed of the model. Experiment results also evaluate the robustness of our architecture even with highly-occluded skeleton poses in the observed prefix sequence. Based on this, we claim that our 2CH-TR stands out as a real-world solution for 3D Human Motion Forecasting in robotics applications. 




\end{document}